\documentclass[9pt]{article}
\usepackage{spconf,amsmath,graphicx}
\usepackage{subcaption}
\usepackage{hyperref}
\usepackage{algorithm}
\usepackage{algorithmic}
\newcommand{\forceindent}{\leavevmode{\parindent=2em\indent}}
\newcommand{\argmax}[1]{\underset{#1}{\operatorname{arg}\,\operatorname{max}}\;}
\newcommand{\argmin}[1]{\underset{#1}{\operatorname{arg}\,\operatorname{min}}\;}
\title{MAN-RECON: MANIFOLD LEARNING FOR RECONSTRUCTION WITH DEEP AUTOENCODER FOR SMART SEISMIC INTERPRETATION}
\name{Ahmad Mustafa and Ghassan AlRegib}
\address{%OLIVES at the Center for Signal and Information Processing,\\
School of Electrical and Computer Engineering,\\
Georgia Institute of Technology, Atlanta, GA, 30332-0250.\\
\{amustafa9, alregib\}@gatech.edu}

\begin{document}
\ninept

\onecolumn

\date{}% Certain latex templates automatically add the compilation date as a footnote in the generated pdf. This command can remove the date in some of the templates but does not work for all.

\onecolumn % make sure you keep this coverpage as one column. In this template, we force the coverpage to be one column with this command and then switch to double column for the remaining of the paper with the \doublecolumn command. 

%\begin{description}[labelindent=-3cm,leftmargin=1cm,style=multiline]
\begin{itemize}
\item[\textbf{Citation}]{Mustafa, Ahmad, and Ghassan AlRegib. "Man-recon: manifold learning for reconstruction with deep autoencoder for smart seismic interpretation." 2021 IEEE International Conference on Image Processing (ICIP). IEEE, 2021.}

\item[\textbf{DOI}]{\url{10.1109/ICIP42928.2021.9506657}}

\item[\textbf{Review}]{Date of presentation: 20th September 2021}

\item[\textbf{Data and Code}]{\url{https://github.com/olivesgatech/Man-Recon}}

\item[\textbf{Bib}] {@inproceedings\{mustafa2021man,\\
  title=\{Man-recon: manifold learning for reconstruction with deep autoencoder for smart seismic interpretation\},\\
  author=\{Mustafa, Ahmad and AlRegib, Ghassan\},\\
  booktitle=\{2021 IEEE International Conference on Image Processing (ICIP)\},\\
  pages=\{2953--2957\},\\
  year=\{2021\},\\
  organization=\{IEEE\}
\}}

% Preprint sharing policy can vary depending on the publisher. Before posting a paper to arXiv, please specifically check the transaction/convference you are targeting. In some transactions, papers are usually added to arxiv after acceptance. Pubslishers usually allow the authors to share accepted version of their papers but not the final formatted version that is provided by the pubisher. In case of sharing preprints, publishers usually ask to add DOI and citation to the paper along with a copyright notice.

\item[\textbf{Contact}]{\href{mailto:amustafa9@gatech.edu}{amustafa9@gatech.edu}  OR \href{mailto:alregib@gatech.edu}{alregib@gatech.edu}\\ \url{https://ghassanalregib.info/} \\ }
\end{itemize}

%Following command sequence was used to start the paper content from the following page and avoid numbering cover page.
\thispagestyle{empty}
\newpage
\clearpage
\setcounter{page}{1}

%Cover page was 1 column. \twocolumn changes the page format back to double column.
\twocolumn

\maketitle

\begin{abstract}
Deep learning can extract rich data representations if provided sufficient quantities of labeled training data. For many tasks however, annotating data has significant costs in terms of time and money, owing to the high standards of subject matter expertise required, for example in medical and geophysical image interpretation tasks. Active Learning can identify the most informative training examples for the interpreter to train, leading to higher efficiency. We  propose an Active learning method based on jointly learning representations for supervised and unsupervised tasks. The learned manifold structure is later utilized to identify informative training samples most dissimilar from the learned manifold from the error profiles on the unsupervised task. We verify the efficiency of the proposed method on a seismic facies segmentation dataset from the Netherlands F3 block survey, significantly outperforming contemporary methods to achieve the highest mean Intersection-Over-Union value of 0.773. 
\end{abstract}
\begin{keywords}
Deep Learning, Active Learning, Autoencoders, Interpretation, Manifolds
\end{keywords}
\section{Introduction and Related Work}
\label{sec:intro}
%General problem description, overview of proposed methodology, contributions.
Deep Learning (DL) has led to ground-breaking advancements in computer vision and image processing research fields, owing to its ability to learn deep hierarchical feature representations directly from the raw image data itself, in contrast to the manual feature engineering approaches required by classical machine learning algorithms. A caveat for learning efficient data representations with DL-based methods is the need for large quantities of labeled training data. Acquiring hand-labeled data in sufficient quantities may not be practical in several applications, such as medical diagnostics, subsurface imaging, microscopy, and many other applications. Such applications require highly qualified experts to carefully interpret images and make decisions about the presence or absence of certain conditions and structures. The complexity of the task increases exponentially when the images have to be labeled in a pixel-wise manner i.e., for segmentation. The time and cost of manual annotation of such data by subject-matter experts is a significant bottleneck in using advanced, deep architectures for automatic interpretation tasks. 

That being said, for most domains, there usually exists a smaller subset of training examples that suffices in creating the representation space for the entire data \cite{pmlr-v80-katharopoulos18a}.  Training the DL model using the identified subset of training instances results in a better generalization performance on unseen test examples than would have been possible otherwise with a similar number of training examples selected arbitrarily. This is in fact what the premise the field of Active Learning is based on \cite{Settles10activelearning}. The human expert, also called the \emph{Oracle}, labels a small number of training examples in the first cycle. In each subsequent cycle thereafter, the method identifies and has the expert label a small set of training examples in the unlabeled dataset likely to add the most information to the machine learning model, subject to a pre-defined criterion e.g., uncertainty \cite{Settles10activelearning}.    

We propose in this work a novel active learning methodology based on learning reconstruction manifolds with Deep Autoencoders. Autoencoders refer to a family of learning models that are trained to reconstruct their inputs. They are designed so that they are only able to reconstruct data sampled from the training distribution, preventing them from regressing to a simple identity mapping. As a useful by-product, they are able to learn the manifold structure of high dimensional data  \cite{MartinezMurcia2020}. \cite{kwon2020backpropagated} utilize such a learned manifold for the task of anomaly detection on image datasets by thresholding the distribution of reconstruction error-based scores on input training examples. 

We view the active learning paradigm as the challenge to identify new training samples most dissimilar from the manifold learnt from preexisting training samples. These training samples are also the ones likely to add the \emph{most} information to the learning model about the dataset that it already doesn't have with preexisting training samples. But there is a caveat: we are majorly interested in supervised, discriminative tasks like classification, segmentation etc. In real life, we would not have ground truth labels for these tasks ahead of time. We could however, make decisions about informative training samples based off the reconstruction manifolds learnt via deep autoencoders. To strengthen the link between the learned manifolds for reconstruction and a supervised tasks like segmentation, \emph{We present a network architecture that simultaneously learns the representations for the two tasks in a joint learning framework.} This way, there is a stronger guarantee that informative training samples identified for the reconstruction task would also apply to the supervised task. We show later that indeed turns out to be the case.

In this paper, we pick one of the most challenging applications and it relates to Earth's subsurface imaging for environmental and energy processes. More specifically, we focus on the task of Earth's subsurface facies segmentation on migrated seismic volumes using. Labeling such data requires resources, time and cost, that cannot be afforded. Thus, we find the setup a good fit for the proposed algorithm. We train an encoder-decoder architecture simultaneously for reconstruction and seismic facies segmentation using the same feature representations within the training phase. In the inference phase, all unlabeled seismic sections/images are scanned and the one with the highest reconstruction error is sampled, labeled, and added to the training dataset for retraining the network for the next cycle. The underlying assumption---based on the shared representation learning framework---is that seismic sections with the highest reconstruction errors are also going to be the ones the network would have performed more poorly on at segmentation. Identifying such training examples would lead to an improved generalization over the whole seismic volume compared to if a similar number of training sections had been sampled randomly or in some other arbitrary fashion. We verify this hypothesis by comparing the proposed work with a baseline active learning method plus two other standard data sampling techniques in the seismic domain. This is the first work of this kind in the domain of subsurface imaging and medical image-based diagnostics, to the best of our knowledge. Due to the limited space, we limit our discussions in this paper to the seismic interpretation application. In short, our major contributions in this work are three fold:
\begin{itemize}
\item Proposing a novel active learning method based on manifold learning with Deep Autoencoders. 
\item Learning informative training samples for segmentation via high reconstruction error sampling of unlabeled sections, made possible by the shared representation learning framework for supervised and unsupervised tasks. 
\item Verifying the efficiency of the proposed work via comparison with other standard sampling approaches in the domain of seismic volume interpretation.
\end{itemize}

\section{Proposed Methodology}
\label{sec:proposedmethod}

\subsection{Dataset}
\label{subsec:dataset}
For the purposes of this work, we use the Netherlands Offshore F3 block's migrated seismic volume and associated interpreted labels as developed by \cite{Alaudah2019}. For convenience, we subsample a slightly smaller volume from the main dataset measuring 400 inlines $\times$ 701 crosslines $\times$ 255 samples in depth. The data contains a total of six facies (rock-types), one assigned to each voxel in the volume. Selected slices from the volume are depicted in Fig.~\ref{fig:setup}. 

\subsection{Architecture}
\label{subsec:architecture}
\begin{figure*}
    \centering
    \includegraphics[width=2\columnwidth]{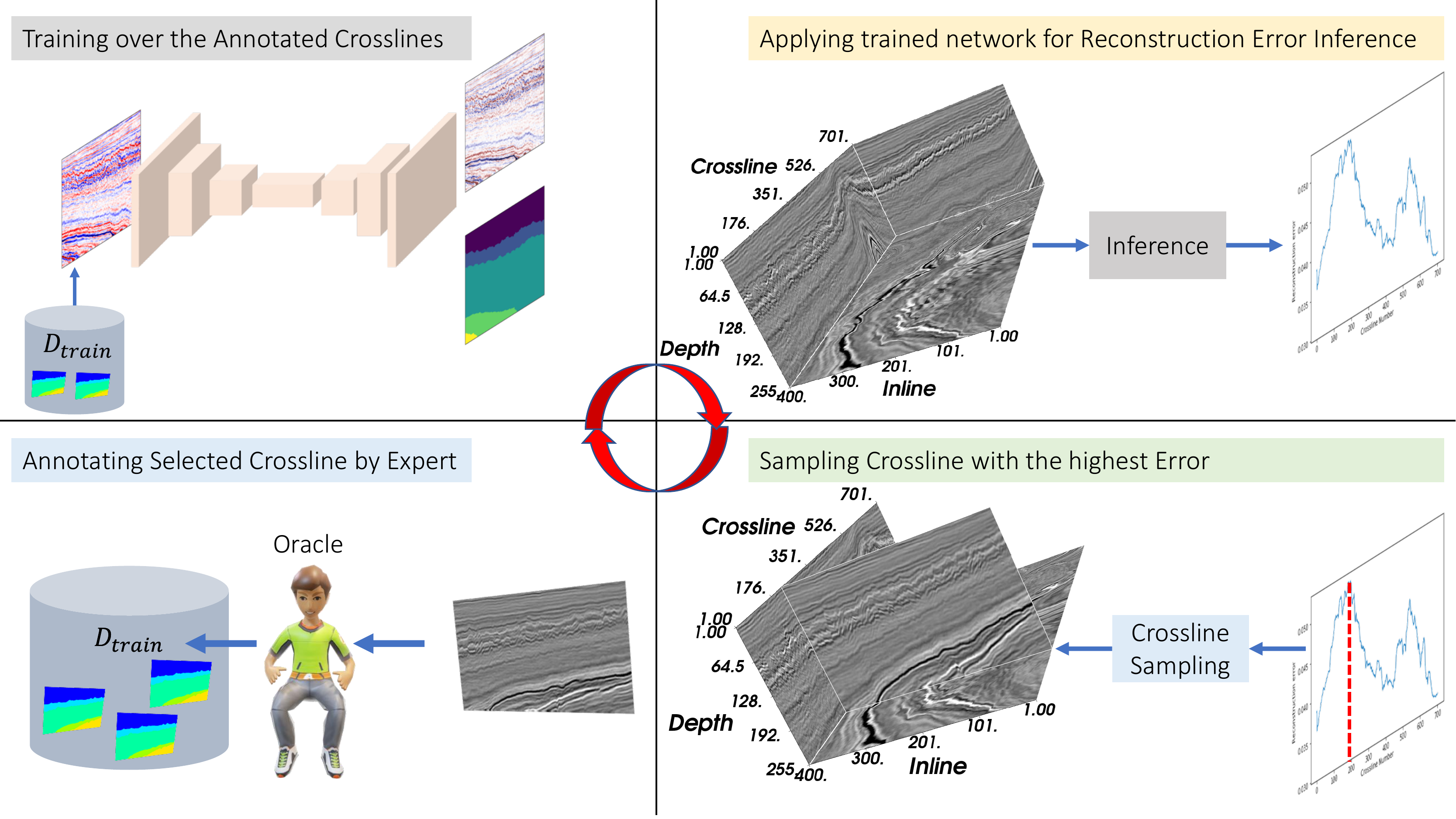}
    \caption{The Active Learning Setup with the proposed methodology.}
    \label{fig:setup}
\end{figure*}

The network architecture used in the work is shown in Fig.~\ref{fig:setup}. It is a convolutional autoencoder with a contracting and expanding path. Each block in the contracting path consists of two convolutional layers followed by a max-pooling layer that downsamples network activations by a factor of two. In contrast, the network blocks in the expanding path each consist of a transposed convolutional layer followed by a regular convolutional layer. The transposed convolutional layer serves to upsample preceding activations by a factor of two. All regular convolutional layers use a kernel size of three with padding to ensure constancy in the dimensions of the activations. Seen overall, it is a very standard image encoder-decoder framework; the encoder successively downsamples input resolution to achieve a more global receptive field while simultaneoulsy increases the number of channels to learn a rich data representation whereas the decoder reverses the downsampling effect of the encoder with a simultaneous reduction in channel size to produce output estimations.   

The input to the network is a complete seismic crossline section measuring 255 samples in depth and 400 samples in width. The network produces as output the reconstructed seismic section and the predicted labels of the various facies present in the section. The two output streams share the same path in the network right up to the penultimate layer, whose output activations are processed by two different 1x1 convolutional layers to estimate the reconstruction of the input seismic section and its facies labels respectively.

\subsection{Training Details}
\label{subsec:trainingdetails}
Fig.~\ref{fig:setup} presents an overview of the active learning setup. The interpreter (oracle) specifies a certain number of active learning cycles, $N$. For the first cycle, the interpreter may randomly select a crossline section in the volume, label the rock facies therein, and provide it to the network for the training phase. The network trains for 300 epochs on the given crossline by minimizing an objective function consisting of the sum of the reconstruction and cross-entropy loss terms as shown below: 

\begin{equation}
    \hat{y}_{j}, \hat{x}_{j} = D_{\Theta}(E_{\Phi}(x_{j})), \quad j=1,\hdots,|\mathcal{D}_{train}|,
    \label{eq:net_pred}
\end{equation}

\begin{equation}
l_{recons} = \frac{1}{|\mathcal{D}_{train}|}\sum_{j=1}^{|\mathcal{D}_{train}|}\left\lVert x_{j}-\hat{x}_{j}\right\rVert^{2}_{2},\text{ and}
\label{eq:recons_loss}
\end{equation}

\begin{equation}
    l_{CE} = \frac{1}{|\mathcal{D}_{train}|}\sum_{j=1}^{|\mathcal{D}_{train}|} CE(y,\hat{y}),
    \label{eq:ce_loss}
\end{equation}
where $\mathcal{D}_{train}$ is the set of all crossline and label pairs in the training set, $D_{\Theta},E_{\Phi}$ are the Decoder and Encoder networks characterized by parameters $\Theta$ and $\Phi$ respectively, $x_{j}$ and $y_{j}$ are the crossline section and label image respectively for j-th training example, and $\hat{x}_{j},\hat{y}_{j}$ are the reconstructed section and predicted label respectively for j-th training example. The optimal model parameters are then solved for by optimizing a weighted sum of the reconstruction and cross-entropy loss terms computed above:

\begin{equation}
\Theta^{*},\Phi^{*} =\argmin{\Theta,\Phi} \alpha\times l_{recons} + \beta\times l_{CE},     
\end{equation}

where $\alpha$ and $\beta$ refer to weights assigned to the reconstruction and cross-entropy losses respectively. The network then enters the inference stage where it predicts the reconstruction for all crossline sections in the unlabeled dataset. The reconstruction losses between the estimations and the original crosslines are computed and stored. The crossline corresponding to the highest loss value is sampled from the dataset, provided to the interpreter for labeling, and then added to the original training dataset. The whole process is then repeated for the specified number of cycles. It is pertinent to mention here that the network is not initialized from scratch in each successive cycle; rather, it is retrained from the point where training last stopped in the previous cycle with the newly sampled and labeled crossline added to the original training dataset. The steps for the complete active learning workflow are listed in Algorithm \ref{alg:activelearning}.

\begin{algorithm}
\caption{Active Learning for Seismic Facies Interpretation}
\label{alg:activelearning}
1. Sample a crossline, $x_{1}$ from $\mathcal{D}_{oracle}=\{x_{j}\}^{M}_{j=1}$.\\
2. Form the initial training dataset, $\mathcal{D}^{0}_{train} = \{(x_{1},y_{1})\}$.\\
3. Initialize the model, $m$.\\
4. For $i=1,\hdots,N$:\\
\forceindent (a) Train $m$ for 300 epochs on $\mathcal{D}^{i-1}_{train}$\\
\forceindent (b) Obtain reconstruction errors for all $x\in  \mathcal{D}_{oracle}$ as\\
\forceindent $\mathcal{E}=\{e_{j}\}^{M}_{j=1}$.\\
\forceindent (c) Sample the crossline with the highest error,\\
\forceindent $x_{s} = \argmax{e_{j}} \mathcal{E}$.\\
\forceindent (d) Form training dataset for next cycle as $\mathcal{D}^{i}_{train}$\\
\forceindent $=\mathcal{D}^{i-1}_{train} \cup \{(x_{s},y_{s})\}.$\\
5. Predict facies labels for all $x\in\mathcal{D}_{oracle}$.
\end{algorithm}

For our purposes, we chose $N=8$, and $\alpha=\beta=1$. The training was carried out using the popular Deep learning library, PyTorch \cite{Pytorch}, with Adam \cite{kingma2017adam} as the optimizer and a learning rate of 0.001. 

\section{Results and Evaluation}
\label{subsec:results}
In addition to the active learning-based data sampling and labeling workflow just described above, we implement three other sampling approaches to compare and contrast the performance of the proposed approach with. They are: 

\begin{itemize}
    \item \textit{Random Sampling}: The workflow as described in Algorithm \ref{alg:activelearning} is implemented, except that a randomly selected crossline section is sampled, labeled, and added to the training dataset at the end of each cycle. 
    \item \textit{Uniform One-sided}: This refers to the case where the interpreter labels the first crossline section in the volume for the first cycle; at each subsequent cycle, a crossline section 20 steps away from the last sampling position is obtained and labeled to be added to the training dataset. 
    \item \textit{Maximum Entropy Sampling}: Entropy is a frequently used metric in active learning settings to identify the training examples with the highest uncertainty. After each cycle, we pick out the crossline with the highest entropy value.
\end{itemize}

To allow for a fair comparison, all parameters are kept constant across the three approaches except for the obvious choice of the sampling criterion. After each cycle, the average segmentation performance over the complete dataset is computed and recorded. The segmentation performance is measured in terms of the mean Intersection-Over-Union (mIOU), also called the Jaccard Index, between the predicted and ground-truth labels. For two finite-size sample sets, $\mathcal{A}$ and $\mathcal{B}$, the Jaccard score is computed as the ratio of the intersection of the two divided by the size of their union, as shown below:
\begin{equation}
    J(\mathcal{A},\mathcal{B}) = \frac{|\mathcal{A}\cap\mathcal{B}|}{|\mathcal{A}\cup\mathcal{B}|}.
\end{equation}
This is especially useful for our case since we are working with a highly imbalanced dataset, where some classes represent less than five percent of the total number of voxels in the volume. mIOU is therefore more likely to be reflective of the true segmentation performance as compared to a metric like pixel accuracy, for example.

\begin{figure}
    \centering
    \includegraphics[width=\columnwidth]{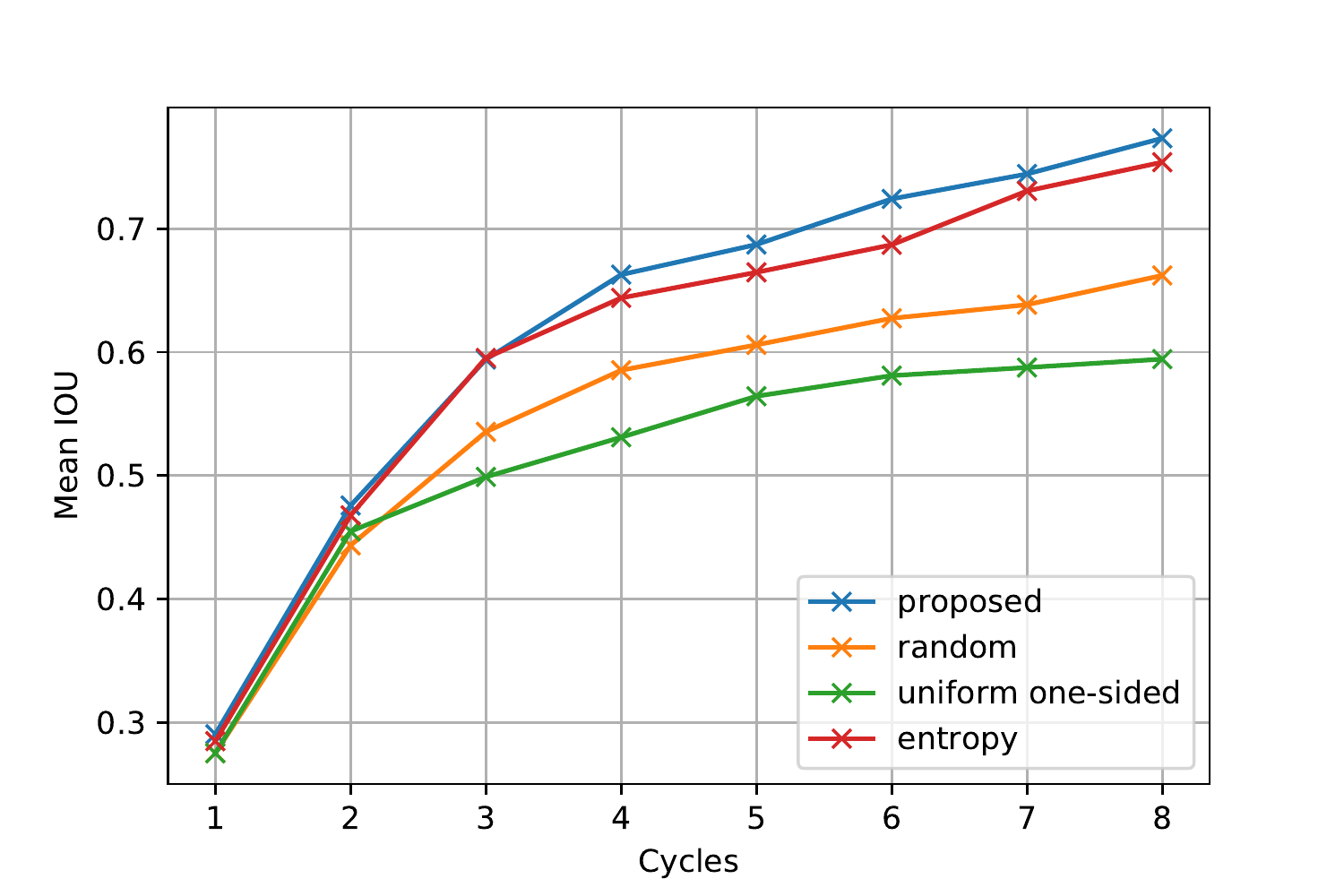}
    \caption{Per Cycle Mean Intersection-Over-Union for the proposed, random, uniform sampling, and entropy-based techniques.}
    \label{fig:results_active_learning}
\end{figure}

The mIOUs per cycle for each of three sampling methodologies are plotted in Fig.~\ref{fig:results_active_learning}. It is quite apparent that the proposed sampling methodology vastly outperforms the random and uniform sampling schemes, achieving close to a ten percentage-point difference over the former. It also beats the baseline entropy-based technique by a smaller margin. Each of the four curves is an average of three trials, lending more credibility to our approach. 

The geological block represented by the seismic volume changes geology as one moves across it in the crossline direction. In some places, the change is slow; in other places, it is faster and more abrupt. To do well overall in terms of facies prediction on the whole volume, the network needs to have a training dataset that well represents the various geological changes throughout the volume. The uniform sampling-based active learning technique performs the worst at the facies segmentation task because the crosslines sampled for training the network all represent the same major geological formation in the volume---it therefore adds no information to the network regarding other formations, leading to a poor segmentation performance over the whole dataset. Random sampling of crosslines attempts to mitigate this shortcoming by not being restricted to sampling from any one region in the volume, but it is still biased towards sampling with a greater likelihood the longer spanning geological formation, potentially missing out on others. The proposed scheme works the best in this regard by sampling the least number of most representative crossline sections in the volume. This is because at each cycle, the network is forced to search for training examples \textit{most dissimilar} to the training set in terms of the reconstruction performance of the learned autoencoder manifold. 

\begin{figure}
\centering
\begin{subfigure}[b]{0.45\textwidth}
   \includegraphics[width=1\linewidth]{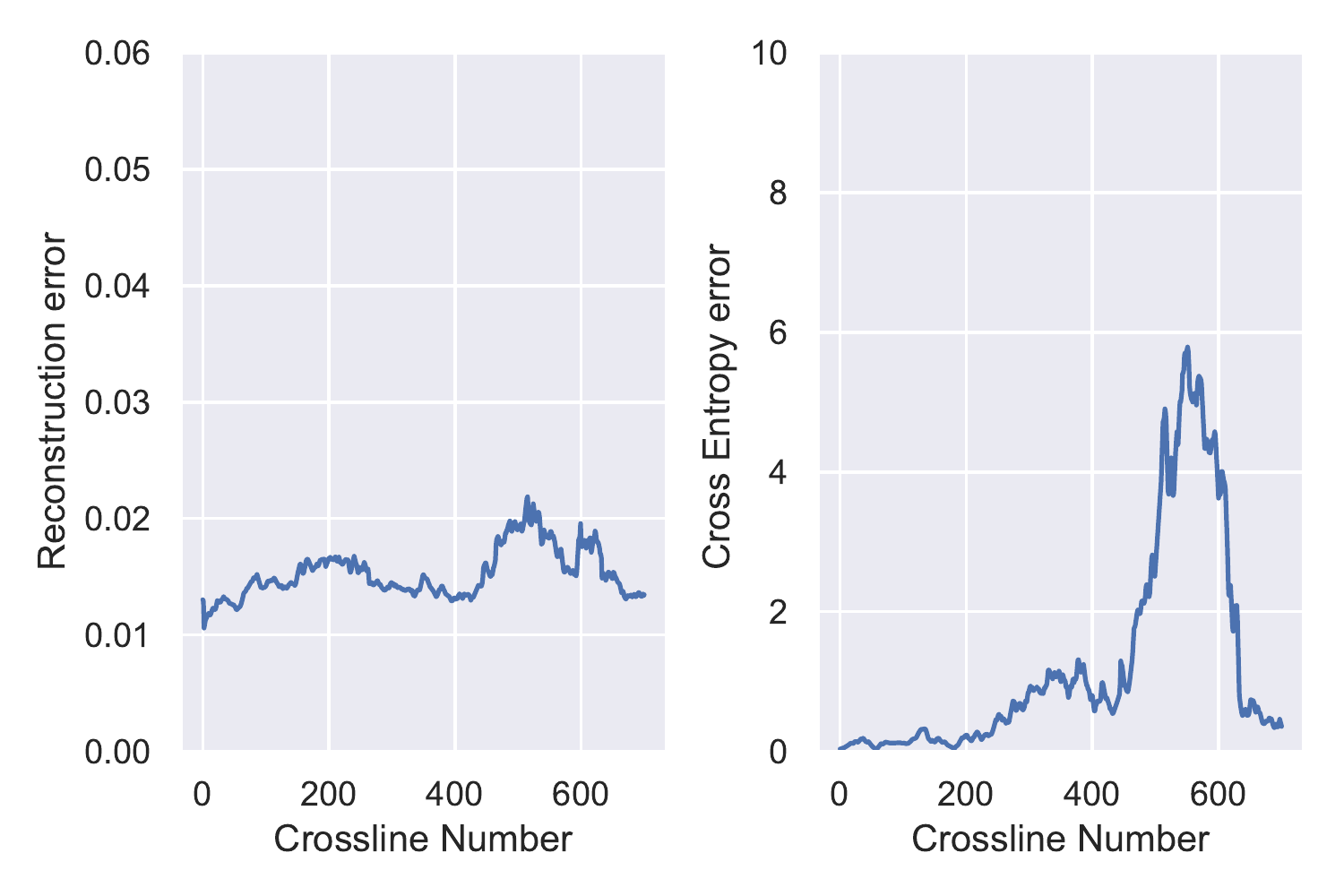}
   \caption{}
   \label{fig:cycle_1} 
\end{subfigure}

\begin{subfigure}[b]{0.45\textwidth}
   \includegraphics[width=1\linewidth]{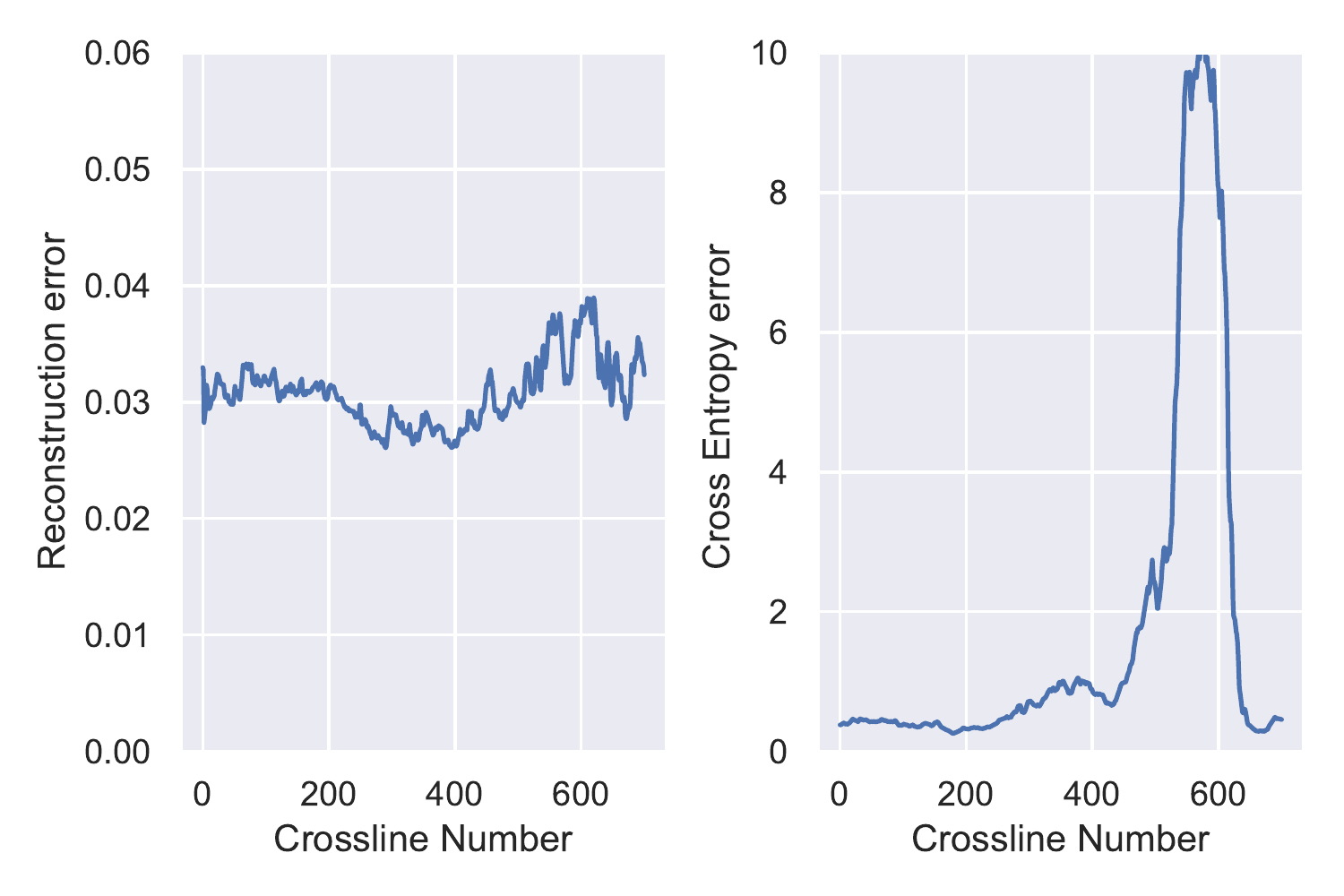}
   \caption{}
   \label{fig:cycle_2}
\end{subfigure}

\caption{Plots of two reconstruction error profiles and their corresponding cross entropy error profiles}
\end{figure}

Finally, Fig.~\ref{fig:cycle_1} and Fig.~\ref{fig:cycle_2} depict the reconstruction error profiles (left) and their corresponding cross-entropy profiles (right) plotted for each of the 701 crosslines in the dataset, for two different active learning cycles. Notice that the latter can only be really observed in the case where we have the ground-truth (true in this case) while the former applies even to the scenario without the ground-truth. It is instructive that the cross-entropy error profile, that ultimately determines the segmentation performance correlates strongly with the reconstruction error profile, as we hoped. Both, for instance, exhibit a peak value around crossline 500 and a trough approximately at crossline position 400. This provides us yet another reassurance that the network tries to learn a similar manifold representation for both the reconstruction and segmentation tasks, so that training samples performing poorly at one task would also be likely to perform poorly at the other.

\section{Conclusion}
\label{Conclusion}
The paper describes an active learning setup to identify the most informative training examples demonstrated with the task of facies interpretation on a migrated seismic dataset. An encoder-decoder architecture is jointly trained for the tasks of reconstruction (unsupervised) and segmentation (supervised) on an initially sampled and labeled crossline section. The network is then used to perform inference on all unlabeled training instances to obtain a reconstruction error profile. The Training example corresponding to the highest error is then sampled, labeled, and added to the training set for the next training cycle. The underlying assumption is that the same training sample would also add the most information about the dataset for the segmentation task due to the shared representation nature of the network. The efficiency of the method is verified by comparing it to other standard sampling approaches in the literature. The proposed work results in the highest mIOU value of 0.773 on the complete dataset over other approaches.

\bibliographystyle{IEEEbib}
\bibliography{refs}

\begin{thebibliography}{1}

\bibitem{pmlr-v80-katharopoulos18a}
Angelos Katharopoulos and Francois Fleuret,
\newblock ``Not all samples are created equal: Deep learning with importance
  sampling,''
\newblock in {\em Proceedings of the 35th International Conference on Machine
  Learning}, Jennifer Dy and Andreas Krause, Eds., Stockholmsmässan, Stockholm
  Sweden, 10--15 Jul 2018, vol.~80 of {\em Proceedings of Machine Learning
  Research}, pp. 2525--2534, PMLR.

\bibitem{Settles10activelearning}
Burr Settles,
\newblock ``Active learning literature survey,''
\newblock Tech. {R}ep., 2010.

\bibitem{MartinezMurcia2020}
Francisco~J. Martinez-Murcia, Andres Ortiz, Juan-Manuel Gorriz, Javier Ramirez,
  and Diego Castillo-Barnes,
\newblock ``Studying the manifold structure of alzheimer{\textquotesingle}s
  disease: A deep learning approach using convolutional autoencoders,''
\newblock {\em {IEEE} Journal of Biomedical and Health Informatics}, vol. 24,
  no. 1, pp. 17--26, Jan. 2020.

\bibitem{kwon2020backpropagated}
Gukyeong Kwon, Mohit Prabhushankar, Dogancan Temel, and Ghassan AlRegib,
\newblock ``Backpropagated gradient representations for anomaly detection,''
\newblock in {\em Proceedings of the European Conference on Computer Vision
  (ECCV)}, 2020.

\bibitem{Alaudah2019}
Yazeed Alaudah, Patrycja Micha{\l}owicz, Motaz Alfarraj, and Ghassan AlRegib,
\newblock ``A machine-learning benchmark for facies classification,''
\newblock {\em Interpretation}, vol. 7, no. 3, pp. SE175--SE187, Aug. 2019.

\bibitem{Pytorch}
Adam Paszke, Sam Gross, Francisco Massa, Adam Lerer, James Bradbury, Gregory
  Chanan, Trevor Killeen, Zeming Lin, Natalia Gimelshein, Luca Antiga, Alban
  Desmaison, Andreas Kopf, Edward Yang, Zachary DeVito, Martin Raison, Alykhan
  Tejani, Sasank Chilamkurthy, Benoit Steiner, Lu~Fang, Junjie Bai, and Soumith
  Chintala,
\newblock ``Pytorch: An imperative style, high-performance deep learning
  library,''
\newblock in {\em Advances in Neural Information Processing Systems 32},
  H.~Wallach, H.~Larochelle, A.~Beygelzimer, F.~d\textquotesingle
  Alch\'{e}-Buc, E.~Fox, and R.~Garnett, Eds., pp. 8024--8035. Curran
  Associates, Inc., 2019.

\bibitem{kingma2017adam}
Diederik~P. Kingma and Jimmy Ba,
\newblock ``Adam: A method for stochastic optimization,'' 2017.

\end{thebibliography}

\end{document}